\documentclass[letterpaper]{article} 
\usepackage{aaai2027}  
\usepackage[hyphens]{url}  
\usepackage{graphicx} 
\usepackage{natbib}  
\usepackage{caption} 
\usepackage{algorithm}
\usepackage{algorithmic}

\usepackage{graphicx}

\usepackage{enumitem}
\usepackage{microtype}
\usepackage{graphicx}
\usepackage{amsmath,amssymb}
\usepackage{booktabs}
\usepackage{multirow}
\usepackage{algorithm}
\usepackage{algorithmic}
\usepackage{cleveref}
\usepackage{subcaption}
\usepackage{lineno}
\usepackage{url}
\usepackage[dvipsnames]{xcolor}
\usepackage[T1]{fontenc}
\usepackage{tcolorbox}
\usepackage{listings}
\usepackage{upquote}
\definecolor{darkblue}{rgb}{0, 0, 0.5}
\definecolor{shupurple}{rgb}{0.5, 0.0, 0.7}
\colorlet{ablationrowhighlight}{ForestGreen!10}
\colorlet{ablationrowhighlight}{Black!8}
\newcommand{\ablationhl}[1]{{\setlength{\fboxsep}{1pt}\colorbox{ablationrowhighlight}{#1}}}
\newcommand{\ablationhlcell}[1]{\multicolumn{1}{@{}c@{}}{\ablationhl{#1}}}
\usepackage{newfloat}
\usepackage{listings}
\DeclareCaptionStyle{ruled}{labelfont=normalfont,labelsep=colon,strut=off} 
\floatstyle{ruled}
\newfloat{listing}{tb}{lst}{}
\floatname{listing}{Listing}

\usepackage{booktabs}

\title{LogicTrack: Auditing Reasoning Trajectories of \\ Large Language Models with Formal Logic Solvers}

\author {
    Jingyu Hu\textsuperscript{\rm 1},
    Shu Yang\textsuperscript{\rm 2},
    Weiru Liu\textsuperscript{\rm 1},
    Di Wang\textsuperscript{\rm 2}
}

\affiliations {
    \textsuperscript{\rm 1}University of Bristol\\
    \textsuperscript{\rm 2}King Abdullah University of Science and Technology\\
}

\begin{document}

\maketitle

\begin{abstract}
Chain-of-Thought (CoT) reasoning has been shown to improve the performance of large language models (LLMs), yet existing optimization methods largely rely on outcome-based feedback, leaving the logical validity of intermediate reasoning steps largely unverified. To address the gap whereby LLMs arrive at correct final answers through logically flawed intermediate reasoning chains, we propose LogicTrack, a neuro-symbolic framework that audits reasoning trajectories by auto-formalizing each reasoning step into symbolic representations and verifying it with automated theorem provers. LogicTrack introduces Solver-Based Backtracking Reward (SBR), a step-wise scoring mechanism that quantifies logical soundness and guides backtracking tree search at inference time.
We further extend LogicTrack to construct supervised fine-tuning (SFT) data with backtracking traces from its trajectories, enabling fine-tuned models to internalize step-wise auditing as an intrinsic capability.
Extensive experiments across 8 reasoning benchmarks and 7 LLMs demonstrate that LogicTrack effectively improves both the verifiability of reasoning chains and final answer pass rate, thereby enhancing overall CoT quality and trustworthiness in high-stakes domains.
\end{abstract}


\section{Introduction}
\label{sec:intro}

Reasoning chains have greatly advanced the capacity of large language models (LLMs) to tackle complex tasks in coding~\citep{gao2023pal}, question answering~\citep{lu2022learn}, and logical reasoning~\citep{symbcot}. Recent models such as OpenAI-o1~\citep{jaech2024openai} and DeepSeek-R1~\citep{guo2025deepseek} further improve performance by encouraging the model to verbalize and extend its thinking trajectory before generating a final answer. However, producing a correct final answer is not sufficient on its own. If the intermediate reasoning steps contain logically invalid operations, users in high-stakes domains such as legal and scientific reasoning cannot trust its conclusions, even when the final answer happens to be correct~\citep{turpin2023language,huang2023towards}. 
This concern is especially important as most current methods use outcome-driven training, which does not guarantee that the reasoning path itself is sound.

This concern has motivated a growing body of research focused on the quality of intermediate reasoning steps. Existing work can be broadly divided into two categories. The first is training-free test-time self-refinement, where the LLM generates a response, receives feedback through a critic model or retrieval mechanism, and applies a search algorithm to regenerate the response based on that feedback \citep{yao2023tree,xie2023self}. The second is training-based process supervision, where process reward models (PRMs) are trained to evaluate the quality of individual reasoning steps, typically using human-annotated or model-generated step-level labels~\citep{lightman2023lets,khalifa2023grace}. However, methods in both categories ultimately rely on human feedback or LLM-based judges to produce feedback and reward signals. Such signals are non-deterministic for logical correctness, prone to inconsistency, and difficult to scale reliably across diverse domains.

\begin{figure*}[t]
\centering
\includegraphics[width=\textwidth]{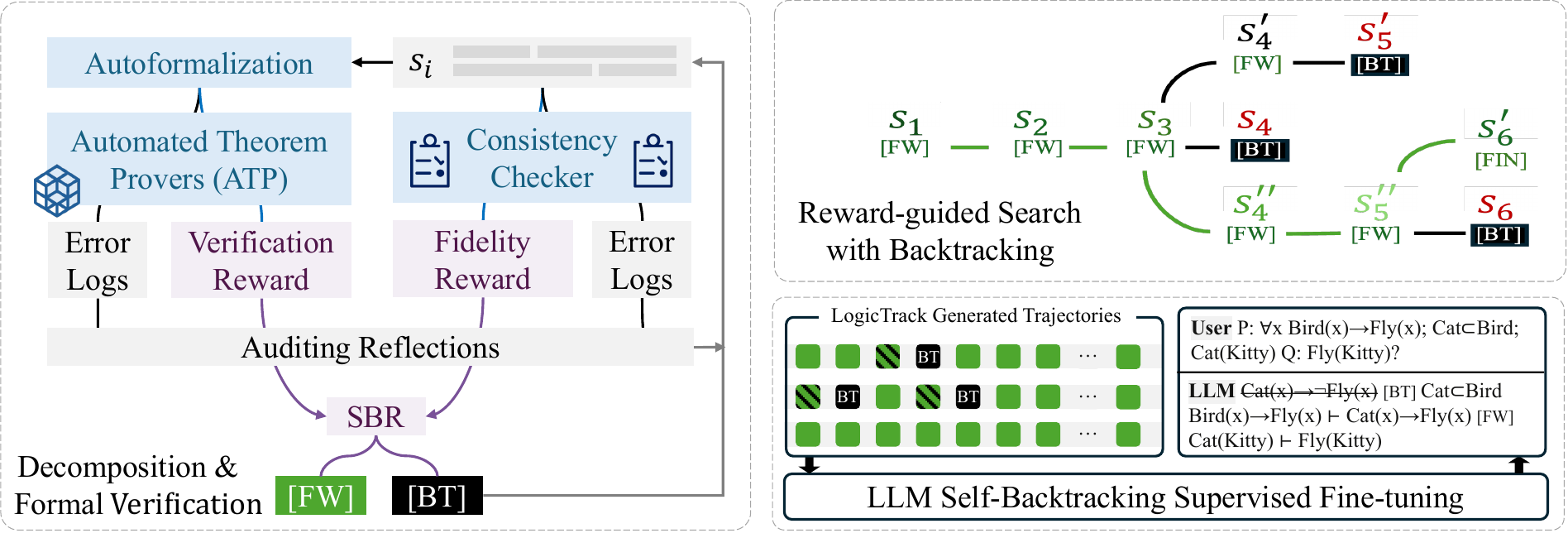}
\caption{Overview of the LogicTrack workflow. Each NL reasoning step is decomposed, formalized into a solver-executable specification, verified, assigned an SBR score, and either forwarded {\small [FW]} to the next step or backtracked {\small [BT]} for regeneration. These generated trajectories can be used for backtracking fine-tuning.}
\label{img:workflow}
\end{figure*}
More recent efforts have turned to neuro-symbolic approaches to make verification more reliable. These include using formal tools for symbolic checking of LLM outputs~\citep{linc111,logic-lm}, auto-formalization of natural language reasoning into solver-verifiable logic, and self-refinement guided by solver feedback~\citep{quan-emnlp24, singh2026verge}. 
Step-wise formal verification for natural language reasoning remains underexplored. Recent LogicReward~\citep{xu2025logicreward} and VeriCoT~\citep{feng2025vericot} represent important initial steps in this direction. Concretely, both methods first generate a complete response and then apply solver-based step verification retroactively, constructing SFT and DPO training data from the verified traces.

However, both methods perform step verification in a post-hoc way. They require a complete reasoning chain and final answer to be generated first, and then audit each step retroactively. This paradigm has two limitations. First, once the full response has been produced, delayed verification raises safety concerns as flawed reasoning may have already been presented to or acted upon by users. Second, a logical error in an intermediate step can propagate through and corrupt subsequent reasoning steps, making after-the-fact correction both more difficult and less reliable.

These limitations naturally raise the question of whether we can \emph{audit LLMs' reasoning trajectories during generation and intervene with timely backtracking to regenerate from the point of failure.} To address this question, we introduce \textbf{LogicTrack}, a neuro-symbolic framework for generation-time auditing and improvement of CoT reasoning through a formally defined verification reward.
Specifically, LogicTrack decomposes each NL reasoning step into premises, explanations, and conclusions, auto-formalizes them into logic specifications, and uses the solver for verification. We define \textbf{Solver-based Backtracking Reward (SBR)}, a fine-grained step-level signal that combines premise fidelity with solver verification. LogicTrack enables a reward-guided backtracking search at inference time. When SBR of a step falls below a predefined threshold, the model triggers a backtrack and regenerates the reasoning from that step onward.

Beyond inference-time auditing, we further extend the application of LogicTrack to \textbf{self-backtracking supervised fine-tuning (SFT)}. Unlike prior self-correction methods that rely on synthetically injected errors to construct backtracking datasets~\citep{yang2025step}, LogicTrack pairs real model-generated flawed trajectories with solver-verified corrections. Fine-tuning on these trajectories with a dedicated backtrack token \texttt{<backtrack>} allows the model to internalize step-wise verification behavior, and we show that this benefit generalizes across multiple benchmarks and models.

Overall, our main contributions are as follows:
\begin{itemize}[itemsep=0pt, topsep=0pt, parsep=0pt, partopsep=0pt]
    \item We propose LogicTrack, a neuro-symbolic framework that audits NL reasoning trajectories via autoformalization and solver verification at inference time. LogicTrack includes a reward-guided backtracking mechanism that uses Solver-based Backtracking Reward to audit and guide model step-wise reasoning.
    \item We evaluate LogicTrack on eight reasoning benchmarks across seven models, showing consistent improvements in both final-answer correctness and intermediate-step validity. Further ablation studies and diverse variants of the search strategy confirm the generalizability of LogicTrack.
    \item We show that LogicTrack can be extended to construct high-quality backtracking SFT data. By introducing a dedicated backtrack token, the fine-tuned model internalizes step-wise verification behavior and achieves self-backtracking in reasoning.
\end{itemize}

\section{Methodology of LogicTrack}
\label{sec:method}

Figure~\ref{img:workflow} illustrates the overall workflow of LogicTrack, which proceeds in three parts: decomposition and formal verification, reward-guided search with backtracking, and the extended application of LogicTrack as higher-quality training data for self-backtracking supervised fine-tuning.

\paragraph{Preliminaries.} When models use natural language to infer logical reasoning tasks, their goal is to determine the logical relationship between a set of given premises $P = \{p_1, \ldots, p_n\}$ and a query $Q$. Each data sample is a triple $x = (P, Q, y)$, where $y \in \mathcal{Y}$ is the ground-truth label from a dataset-specific label set $\mathcal{Y}$. Although label names (e.g., \textit{entailment/contradiction}, \textit{yes/no/uncertain}) differ across datasets, each $\mathcal{Y}$ corresponds to one of three underlying logical relations: $P \models Q$, $P \models \neg Q$, or neither.
Given a sample $x$, the model response is an ordered sequence of $T$ reasoning steps $(s_1, s_2, \ldots, s_T)$ followed by a final prediction $\hat{y}$. To ensure the quality of both the final prediction and the intermediate reasoning trajectories, we propose LogicTrack to audit the logical validity of each intermediate step and trigger backtracking when necessary.

\subsection{Decomposition and Formal Verification}
\label{sec:verification}

\paragraph{Step Decomposition}
\label{sec:formalization}

We specify in the LLM response format that each reasoning step needs to present new conclusion(s) derived from context $C_i$ and explanation $E_i$, and the details are explained below.
Let $\mathcal{K}_{i} = \{q_1, \ldots, q_{i}\}$ denote the set of derived conclusions from the first $i$ verified reasoning steps ($\mathcal{K}_0 = \emptyset$). Specifically, the step $s_i$ can be decomposed into three components $s_i = (C_i, E_i, q_i)$, defined as below.
With both $C_i$ and $E_i$ stated explicitly, the conclusion $q_i$ should follow by logical entailment alone.
The verified $q_i$ is then incorporated into $\mathcal{K}_{i}$, making it available as context for subsequent steps.

\begin{itemize}[itemsep=0pt, topsep=0pt, parsep=0pt, partopsep=0pt]
    \item \textbf{Context} $C_i$: the premises from $P$ and prior verified conclusions relevant to this step, i.e., $C_i \subseteq P \cup \mathcal{K}_{i-1}$.
    \item \textbf{Explanation} $E_i$: implicit assumptions and background knowledge that the step relies on but does not state, e.g., ``a father is a parent''.
    \item \textbf{Conclusion} $q_i$: the new claim derived from $C_i$ and $E_i$.
\end{itemize}

\paragraph{Autoformalization and Verification}
\label{sec:formalization_verification}

Unlike post-hoc verification methods that check the entire reasoning chain after the response is fully generated, LogicTrack performs verification during reasoning in a step-wise autoregressive way.
Starting from the $i{-}1$ verified steps $(s_1, \ldots, s_{i-1})$ and knowledge base $\mathcal{K}_{i-1}$, the reasoning LLM generates the next candidate step $s_i$, conditioned on $(P, Q, s_1, \ldots, s_{i-1})$.
The candidate is then decomposed, formalized and verified by the logic solver.

For each candidate step $s_i$, the formalization model receives components $(C_i, E_i, q_i)$ and formalizes them into an executable logic specification $F_i$ in SMT-LIB format.
The context and explanations are jointly formalized as a conjunction of solver assertions $\Phi_i$, i.e., $\Phi_i$ encodes $C_i \cup E_i$. The conclusion $q_i$ is encoded as the proof goal, yielding $F_i = (\Phi_i,\, q_i)$, so that verification reduces to checking if $\Phi_i \models q_i$.
Given $F_i = (\Phi_i,\, q_i)$, LogicTrack verifies whether the candidate step is well-formed and logically justified.
The solver first checks syntactic validity of the generated formalization; if the specification is malformed, the autoformalization module is re-invoked up to $M$ times.
Once a valid specification is obtained, the solver tests whether $\Phi_i \models q_i$ via proof by refutation, that is, it checks whether $\Phi_i \cup \{\neg q_i\}$ is unsatisfiable, and also records diagnostic signals such as contradiction or undecidability.
These verification outcomes, together with the judge's assessment of explanation grounding, are stored as a step-level audit log and converted into the scalar reward used by the search controller.

\subsection{Reward-Guided Backtracking Search}
\label{sec:tree_search}

We consider two aspects of effective reasoning. The model must ground its reasoning in the given premises, and each inference must be logically sound. We design a reward function called the Solver-based Backtracking Reward (SBR) to quantify both aspects.
LogicTrack assigns each step $s_i$ a composite reward 
$\mathrm{SBR}(s_i) = \lambda \cdot R^{\mathrm{fidelity}}(\Phi_i;\, P) + (1-\lambda) \cdot R^{\mathrm{verify}}(F_i)$.

\begin{itemize}[itemsep=0pt, topsep=0pt, parsep=0pt, partopsep=0pt]
    \item \textbf{Fidelity reward} $R^{\mathrm{fidelity}}(\Phi_i;\, P)$: Checks the consistency of $\Phi_i$, including if each assertion is grounded in the original premises $P$ (not sourced from the query $Q$), if introduced commonsense is acceptable, and if $\Phi_i$ is jointly satisfiable. We define $R^{\mathrm{fidelity}}(\Phi_i;\, P) = 1$ if all checks pass, and $0$ otherwise.
    \item \textbf{Verification reward} $R^{\mathrm{Verify}}(F_i)$: The verification score is set to $1$ if the specification $F_i$ is syntactically correct and the step is logically valid ($\Phi_i \models q_i$).
If $F_i$ is syntactically invalid, autoformalization will be regenerated up to $M$ times before the step is marked as unverifiable and $R^{\mathrm{Verify}}(F_i) = 0$.
We also set $R^{\mathrm{Verify}}(F_i) = 0$ if the solver actively refutes the step ($\Phi_i \models \neg q_i$).
For undecided steps, the verifier assigns partial credit $R^{\mathrm{Verify}}(F_i) = \alpha$, where $\alpha \in (0,1)$ is a discounting constant, since these steps may simply exceed the solver's decision procedure without being logically wrong.
\end{itemize}
LogicTrack builds the verified reasoning chain via reward-guided backtracking search to decide whether to forward or backtrack on each candidate step.
At each step $t$, the search maintains a verified prefix $(s_1, \ldots, s_{t-1})$ and evaluates a new candidate $s_t$ against two quality criteria: the step-level reward $\mathrm{SBR}(s_t)$ and running mean over all verified steps $\overline{\mathrm{SBR}}(t) = \frac{1}{t} \sum_{i=1}^{t} \mathrm{SBR}(s_i)$.
The search starts from an empty prefix and proceeds one step at a time.
A candidate $s_t$ is forwarded if $\mathrm{SBR}(s_t) \geq \theta_{\mathrm{step}} \quad \text{and} \quad \overline{\mathrm{SBR}}(t) \geq \theta_{\mathrm{avg}}$.
In that case, $s_t$ is appended to the verified prefix and $\mathcal{K}_{t-1}$ is updated to $\mathcal{K}_t$.

When triggering backtracking, LogicTrack collects the audit logs for $s_t$ and re-prompts the model with the verified prefix $(s_1, \ldots, s_{t-1})$ along with concise failure feedback derived from error logs. The model then regenerates from position $t$ onward. This backtracking process repeats until a viable solution is found. We provide implementation details and discussion of other backtracking strategies like Beam search and MCTS in Appendix A.1~\ref{apx:backtrack-variant}.

\subsection{LogicTrack with Backtracking-SFT}
\label{sec:sft_results}

\begin{figure}[t]
    \centering
    \includegraphics[width=1\linewidth]{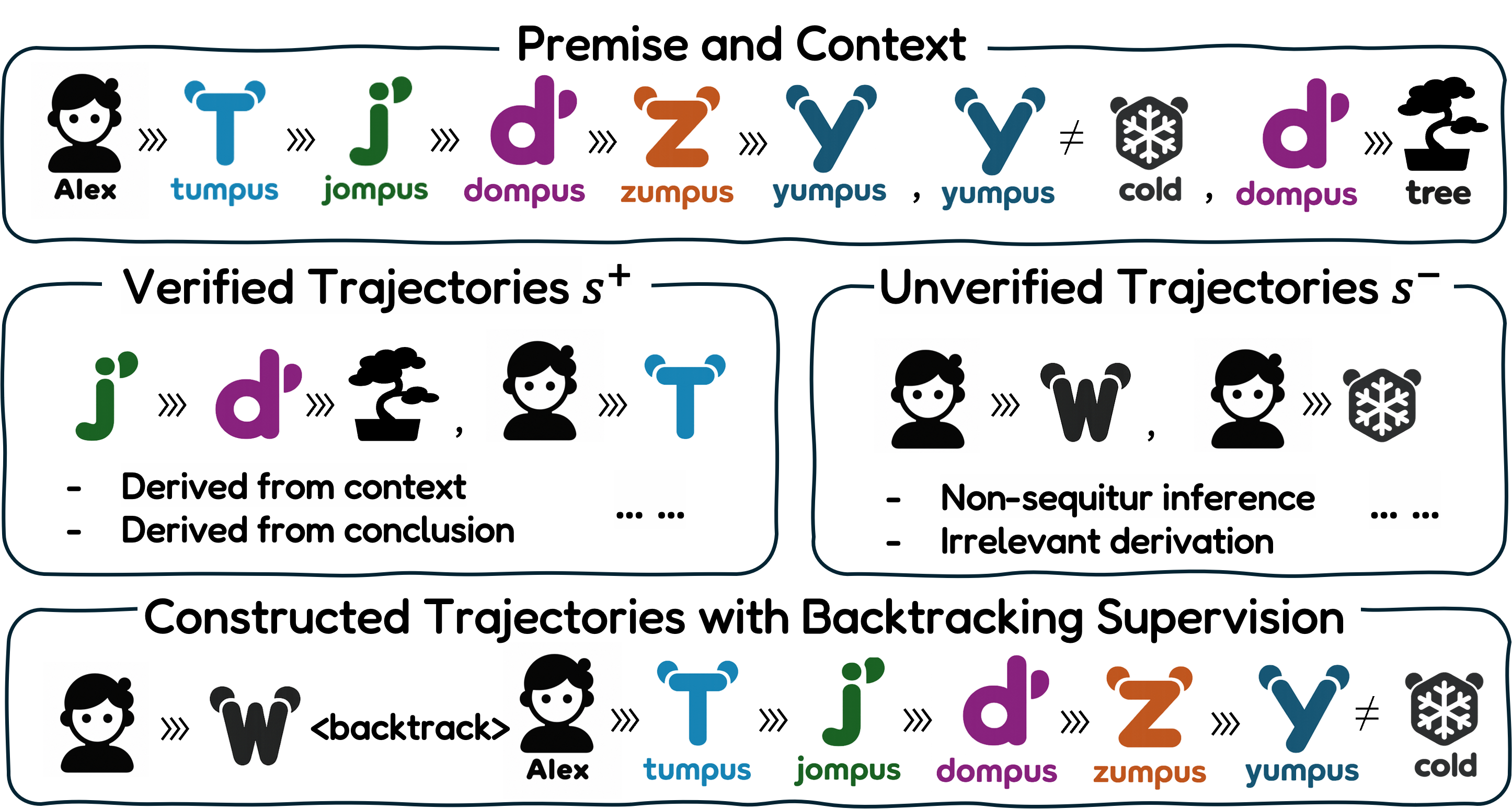}
    \caption{An example of LogicTrack-SFT training data construction with backtracking trajectories.}
    \label{fig:case-study-backtracking-trajectories}
\end{figure}

We introduce LogicTrack-SFT, an extension of LogicTrack that converts its backtracking trajectories into supervised fine-tuning data, allowing the model to internalize step-level auditing as an intrinsic capability.

Recent work has shown that training on explicit error-correction trajectories can teach models to revise flawed reasoning. An et al.~\citep{an2023lema} construct mistake-correction pairs; Ye et al.~\citep{ye2024learnmistakes} show that retry data with erroneous steps immediately followed by corrections improves reasoning; and Self-Backtracking~\citep{yang2025selfbacktracking} further introduces a dedicated backtracking token to learn when and where to revise a partial trajectory. 
However, the supervision signals in these methods are largely synthesized from artificially injected or sampled errors in mathematical settings, where correctness is straightforward to verify.
We build on the same backtracking intuition but focus on a harder setting: using LogicTrack to construct verifier-grounded backtracking steps for NL logic reasoning.

Specifically, SFT training data is constructed from LogicTrack's backtracking trajectories.
We treat the verified correct path from the search tree as $s^+$ and the backtracked steps that triggered backtracking events as $s^-$.
A training trajectory with backtrack at position $k$ is assembled by concatenating the correct prefixes, the backtracked step $s_k^{-}$, a special backtrack token, and the verified correct continuation:
$ \{s_1^{+} \cdots s_{k-1}^{+} \;\; s_k^{-} \;\; \texttt{<backtrack>} \;\; s_k^{+} \cdots s_n^{+} \}$. For example, one possible training trajectory with backtracking in the workflow demo in Figure \ref{img:workflow} is $\{ s_1, s_2, s_3, s_{4}, \texttt{<backtrack>}, s''_{4}, s''_5, s'_6 \}$. 
Figure \ref{fig:case-study-backtracking-trajectories} presents a more intuitive example, where we aim to determine whether Alex is cold based on the given premises, such as `Alex is Tumpus' and `dompus is zumpus'.
We consider both single-round and multi-round backtracking during the synthetic stage, resulting in a set of $\sim$6{,}000 samples for training. 
Given the constructed trajectory $\mathbf{s}$, LogicTrack-SFT minimizes the standard autoregressive negative log-likelihood
$\mathcal{L}_{\mathrm{SFT}}(\theta)
=
-\sum_{t=1}^{n}
\log p_{\theta}
\left(
s_t \mid P,Q,s_{<t}
\right)$. The backtrack token \texttt{<backtrack>} is added to the model vocabulary and trained end-to-end, teaching the model to recognize flawed steps and self-correct without external critics. The detailed implementations can be found in Appendix A.2\ref{apx:backtrack-sft}.

\section{Experiments and Results}
\label{sec:experiments}

This section presents the experimental setup and the discussions of overall performance, ablation studies, and LogicTrack's extended application to SFT self-backtracking.

\colorlet{logictrackshade}{white}
\colorlet{logictrackshade}{black!7}
\newcommand{\logictracktitle}{\colorbox{logictrackshade}{LogicTrack (Ours)}}
\newcommand{\logictrackcell}[1]{%
  \begingroup
  \setlength{\fboxsep}{1pt}%
  \smash{\colorbox{logictrackshade}{#1}}%
  \endgroup
}
\newcommand{\resultpair}[2]{& #1 & \logictrackcell{#2}}

\begin{table*}[t]
\centering
\scriptsize
\setlength{\tabcolsep}{3pt}
\resizebox{\textwidth}{!}{
\begin{tabular}{@{}ll*{14}{c}@{}}
\hline
\textbf{Dataset} & \textbf{Metric} & \multicolumn{2}{c}{\textbf{Gemini-2.5}} & \multicolumn{2}{c}{\textbf{GPT-4o-mini}} & \multicolumn{2}{c}{\textbf{GPT-5-nano}} & \multicolumn{2}{c}{\textbf{Llama-3.1-8B}} & \multicolumn{2}{c}{\textbf{Mistral-7B}} & \multicolumn{2}{c}{\textbf{Qwen2.5-7B}} & \multicolumn{2}{c}{\textbf{Qwen2.5-14B}} \\
\cmidrule(lr){3-4}\cmidrule(lr){5-6}\cmidrule(lr){7-8}\cmidrule(lr){9-10}\cmidrule(lr){11-12}\cmidrule(lr){13-14}\cmidrule(lr){15-16}
 & \resultpair{\textbf{BASE}}{\textbf{Ours}} \resultpair{\textbf{BASE}}{\textbf{Ours}} \resultpair{\textbf{BASE}}{\textbf{Ours}} \resultpair{\textbf{BASE}}{\textbf{Ours}} \resultpair{\textbf{BASE}}{\textbf{Ours}} \resultpair{\textbf{BASE}}{\textbf{Ours}} \resultpair{\textbf{BASE}}{\textbf{Ours}} \\
\hline
\multirow{3}{*}{FOLIO} & \textbf{VWA} $\Uparrow$ \resultpair{21.17}{\textbf{56.37}} \resultpair{46.35}{\textbf{59.91}} \resultpair{46.35}{\textbf{51.36}} \resultpair{27.76}{\textbf{39.40}} \resultpair{32.87}{\textbf{43.82}} \resultpair{31.90}{\textbf{48.97}} \resultpair{45.14}{\textbf{57.53}} \\
 & \textbf{OA} $\Uparrow$ \resultpair{79.31}{77.59} \resultpair{68.97}{\textbf{74.14}} \resultpair{71.70}{\textbf{73.28}} \resultpair{48.28}{45.69} \resultpair{48.28}{\textbf{54.31}} \resultpair{59.48}{\textbf{65.52}} \resultpair{75.00}{\textbf{75.00}} \\
 & \textbf{UR} $\Downarrow$ \resultpair{75.63}{\textbf{27.79}} \resultpair{34.74}{\textbf{19.76}} \resultpair{38.75}{\textbf{31.90}} \resultpair{41.85}{\textbf{17.45}} \resultpair{33.75}{\textbf{26.72}} \resultpair{47.65}{\textbf{28.41}} \resultpair{42.09}{\textbf{28.00}} \\
\hline
\multirow{3}{*}{ESNLI} & \textbf{VWA} $\Uparrow$ \resultpair{31.09}{\textbf{39.08}} \resultpair{46.38}{\textbf{87.21}} \resultpair{30.10}{\textbf{44.32}} \resultpair{11.89}{\textbf{19.07}} \resultpair{16.12}{\textbf{22.74}} \resultpair{35.25}{\textbf{50.52}} \resultpair{39.50}{\textbf{53.89}} \\
 & \textbf{OA} $\Uparrow$ \resultpair{66.67}{\textbf{66.67}} \resultpair{86.05}{\textbf{100.00}} \resultpair{65.12}{63.57} \resultpair{22.48}{21.71} \resultpair{31.01}{\textbf{31.01}} \resultpair{60.47}{\textbf{64.34}} \resultpair{68.22}{65.89} \\
 & \textbf{UR} $\Downarrow$ \resultpair{52.84}{\textbf{42.96}} \resultpair{45.87}{\textbf{12.79}} \resultpair{53.36}{\textbf{29.46}} \resultpair{38.45}{\textbf{11.18}} \resultpair{35.84}{\textbf{21.64}} \resultpair{40.71}{\textbf{23.58}} \resultpair{40.61}{\textbf{15.67}} \\
\hline
\multirow{3}{*}{LogiQA} & \textbf{VWA} $\Uparrow$ \resultpair{24.91}{20.36} \resultpair{34.33}{\textbf{44.98}} \resultpair{24.30}{\textbf{40.23}} \resultpair{25.97}{\textbf{44.72}} \resultpair{28.11}{\textbf{36.27}} \resultpair{24.81}{\textbf{40.38}} \resultpair{29.17}{\textbf{46.70}} \\
 & \textbf{OA} $\Uparrow$ \resultpair{62.50}{\textbf{65.91}} \resultpair{61.93}{54.55} \resultpair{50.86}{\textbf{55.11}} \resultpair{46.02}{\textbf{50.00}} \resultpair{44.32}{\textbf{46.02}} \resultpair{51.70}{\textbf{51.70}} \resultpair{55.11}{\textbf{62.50}} \\
 & \textbf{UR} $\Downarrow$ \resultpair{63.07}{68.56} \resultpair{46.26}{\textbf{16.05}} \resultpair{54.82}{\textbf{31.70}} \resultpair{42.25}{\textbf{12.47}} \resultpair{39.62}{\textbf{23.48}} \resultpair{50.17}{\textbf{25.20}} \resultpair{50.43}{\textbf{27.22}} \\
\hline
\multirow{3}{*}{Pron.QA} & \textbf{VWA} $\Uparrow$ \resultpair{35.71}{\textbf{52.43}} \resultpair{78.70}{\textbf{81.29}} \resultpair{88.83}{\textbf{89.76}} \resultpair{56.64}{\textbf{71.52}} \resultpair{57.01}{\textbf{62.60}} \resultpair{60.02}{\textbf{82.27}} \resultpair{71.67}{\textbf{72.81}} \\
 & \textbf{OA} $\Uparrow$ \resultpair{99.00}{\textbf{99.00}} \resultpair{90.00}{\textbf{94.00}} \resultpair{100.00}{\textbf{100.00}} \resultpair{85.00}{\textbf{87.00}} \resultpair{68.00}{\textbf{73.00}} \resultpair{92.00}{\textbf{97.00}} \resultpair{96.00}{\textbf{97.00}} \\
 & \textbf{UR} $\Downarrow$ \resultpair{64.29}{\textbf{46.69}} \resultpair{15.05}{\textbf{14.28}} \resultpair{11.17}{\textbf{10.24}} \resultpair{34.17}{\textbf{20.46}} \resultpair{18.89}{\textbf{16.29}} \resultpair{35.14}{\textbf{15.73}} \resultpair{25.44}{\textbf{24.82}} \\
\hline
\multirow{3}{*}{PW} & \textbf{VWA} $\Uparrow$ \resultpair{32.45}{\textbf{53.61}} \resultpair{40.21}{\textbf{63.82}} \resultpair{75.41}{\textbf{78.02}} \resultpair{36.65}{\textbf{42.91}} \resultpair{19.33}{\textbf{29.02}} \resultpair{41.95}{\textbf{53.58}} \resultpair{48.35}{\textbf{49.61}} \\
 & \textbf{OA} $\Uparrow$ \resultpair{88.44}{86.43} \resultpair{54.27}{\textbf{79.90}} \resultpair{100.00}{99.50} \resultpair{59.80}{57.29} \resultpair{27.14}{\textbf{39.20}} \resultpair{63.32}{\textbf{66.83}} \resultpair{80.40}{78.39} \\
 & \textbf{UR} $\Downarrow$ \resultpair{65.13}{\textbf{36.97}} \resultpair{37.45}{\textbf{23.82}} \resultpair{24.59}{\textbf{21.47}} \resultpair{40.77}{\textbf{29.10}} \resultpair{38.31}{\textbf{26.83}} \resultpair{38.19}{\textbf{22.35}} \resultpair{43.59}{\textbf{38.15}} \\
\hline
\multirow{3}{*}{QASC} & \textbf{VWA} $\Uparrow$ \resultpair{20.78}{\textbf{55.88}} \resultpair{54.43}{\textbf{71.35}} \resultpair{30.82}{\textbf{43.04}} \resultpair{47.89}{\textbf{64.02}} \resultpair{25.67}{\textbf{40.64}} \resultpair{0.00}{\textbf{0.00}} \resultpair{59.09}{\textbf{80.59}} \\
 & \textbf{OA} $\Uparrow$ \resultpair{86.30}{\textbf{91.78}} \resultpair{93.15}{90.41} \resultpair{61.64}{\textbf{63.01}} \resultpair{84.93}{75.34} \resultpair{53.42}{\textbf{60.27}} \resultpair{87.67}{\textbf{87.67}} \resultpair{91.78}{\textbf{94.52}} \\
 & \textbf{UR} $\Downarrow$ \resultpair{76.14}{\textbf{44.12}} \resultpair{42.60}{\textbf{23.29}} \resultpair{42.58}{\textbf{26.71}} \resultpair{44.65}{\textbf{16.30}} \resultpair{56.06}{\textbf{33.56}} \resultpair{100.00}{\textbf{100.00}} \resultpair{37.95}{\textbf{16.67}} \\
\hline
\multirow{3}{*}{SemEval} & \textbf{VWA} $\Uparrow$ \resultpair{32.20}{25.04} \resultpair{48.38}{\textbf{56.27}} \resultpair{33.30}{\textbf{34.14}} \resultpair{30.53}{\textbf{42.93}} \resultpair{37.68}{\textbf{40.39}} \resultpair{38.20}{\textbf{48.82}} \resultpair{42.61}{\textbf{50.28}} \\
 & \textbf{OA} $\Uparrow$ \resultpair{93.68}{\textbf{93.68}} \resultpair{82.63}{82.11} \resultpair{50.79}{46.84} \resultpair{60.53}{\textbf{60.53}} \resultpair{65.79}{64.74} \resultpair{74.74}{\textbf{80.53}} \resultpair{81.58}{80.00} \\
 & \textbf{UR} $\Downarrow$ \resultpair{63.77}{73.98} \resultpair{40.89}{\textbf{32.89}} \resultpair{40.44}{\textbf{33.84}} \resultpair{50.91}{\textbf{32.15}} \resultpair{43.29}{\textbf{36.18}} \resultpair{52.10}{\textbf{41.01}} \resultpair{48.35}{\textbf{36.44}} \\
\hline
\multirow{3}{*}{SARA} & \textbf{VWA} $\Uparrow$ \resultpair{36.87}{23.67} \resultpair{40.51}{\textbf{65.90}} \resultpair{39.89}{\textbf{54.46}} \resultpair{32.13}{\textbf{57.98}} \resultpair{38.81}{\textbf{49.39}} \resultpair{36.00}{\textbf{53.52}} \resultpair{42.67}{\textbf{57.00}} \\
 & \textbf{OA} $\Uparrow$ \resultpair{77.94}{77.21} \resultpair{60.29}{\textbf{73.16}} \resultpair{59.93}{\textbf{68.75}} \resultpair{52.94}{\textbf{63.60}} \resultpair{59.93}{58.82} \resultpair{62.50}{\textbf{69.49}} \resultpair{65.81}{\textbf{68.38}} \\
 & \textbf{UR} $\Downarrow$ \resultpair{55.04}{68.36} \resultpair{32.55}{\textbf{10.81}} \resultpair{37.67}{\textbf{20.20}} \resultpair{39.10}{\textbf{9.89}} \resultpair{37.73}{\textbf{17.51}} \resultpair{42.40}{\textbf{24.34}} \resultpair{39.33}{\textbf{20.02}} \\
\hline
\end{tabular}
}
\caption{Benchmark-level Performance Comparison of VWA\%, OA\%, and UR\% between BASE and LogicTrack (Ours) across Eight Reasoning Data Benchmarks and Seven Models. }
\label{tab:main_results}
\end{table*}

\subsection{Experimental Setup}
\label{sec:exp_setup}

\paragraph{Datasets and Baselines.}
We evaluate on a diverse collection of natural-language reasoning benchmarks including e-SNLI~\citep{camburu2018esnli}, FOLIO~\citep{han2024folio}, LogiQA~\citep{liu2020logiqa}, ProntoQA~\citep{prontoqaRef}, ProofWriter~\citep{tafjord2021proofwriter}, QASC~\citep{khot2020qasc}, SARA~\citep{holzenberger2020dataset}, and SemEval~\citep{semeval2026task11}.

Each dataset is split into train and test set following \cite{xu2025logicreward}.
The evaluation covers seven large language models with varying architectures and scales, including three proprietary models (Gemini-2.5~\citep{comanici2025gemini}, GPT-4o~\citep{hurst2024gpt4o}, and GPT-5~\citep{openai2025gpt5}) and four open-weight models (Llama~\citep{llama3}, Mistral~\citep{jiang2023mistral7b}, Qwen2.5-7B, and Qwen2.5-14B \citep{yang2024qwen2}). We use the vanilla CoT as our primary baseline (BASE).
Additional baseline comparisons (i.e. LogicReward) can be found in Appendix C, and more detailed descriptions of the datasets and models are reported in Appendix B.
\paragraph{Metrics.}
We evaluate models using four metrics that assess outcome-level performance, process-level performance, and their combination.
\textbf{Outcome Accuracy (OA)} is the fraction of LLM responses where the final answer is correct.
\textbf{Verified Ratio (VR)} measures the average proportion of reasoning steps that pass solver checks.
\textbf{Unverified Ratio (UR)} is the complementary fraction of unverified steps.
\textbf{Verified Utility (VU)} refers to answer accuracy restricted to all fully verified steps. 
\textbf{Verification-Weighted Accuracy (VWA)} combines both aspects by weighting each correct prediction by its per-problem verification ratio.
More detailed metric descriptions are provided in Appendix B.

\subsection{Main Results: Higher Outcome Correctness and Step Verifiability}
\label{sec:overall_performance}

\begin{figure}[!t]
\centering
\includegraphics[width=1\columnwidth,trim=0 2 0 10,clip]{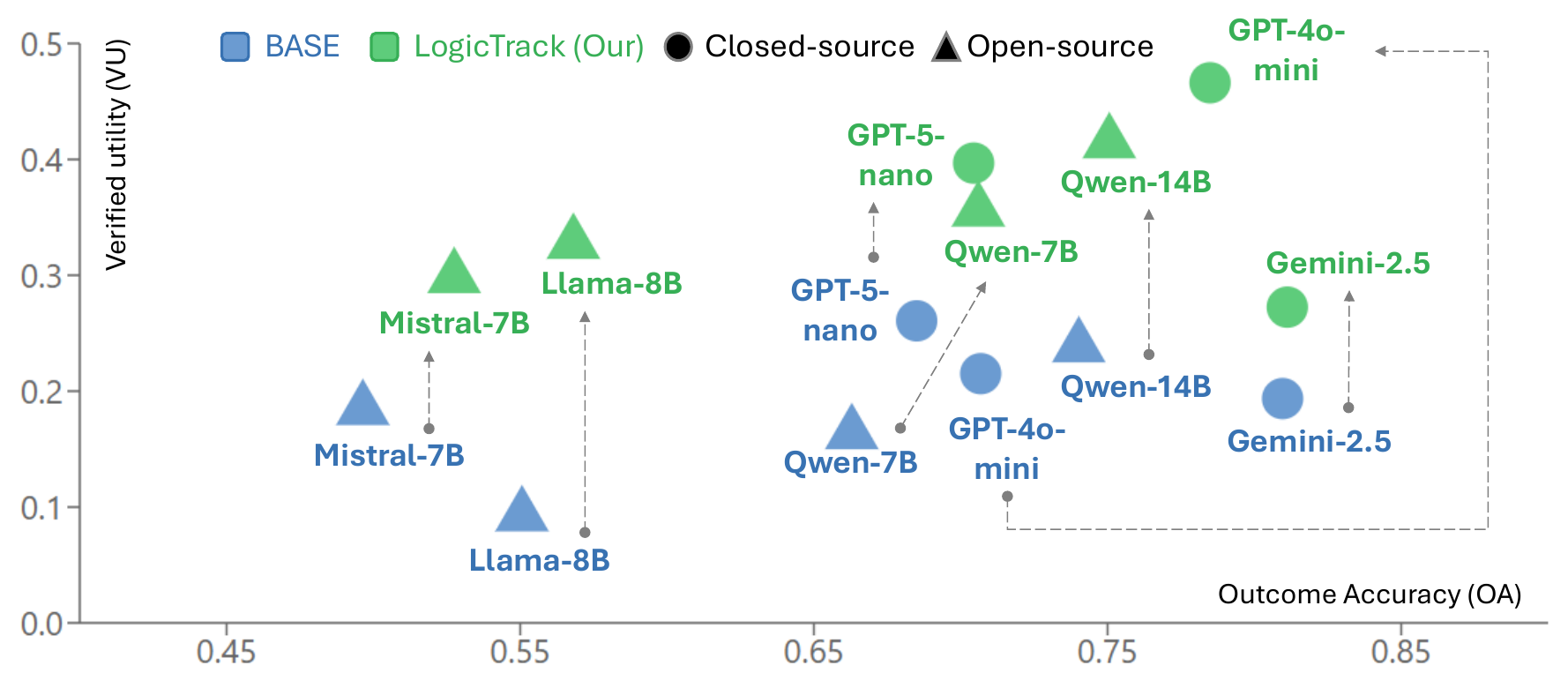}
\caption{Overall OA and VU for all models averaged across eight benchmarks under BASE and LogicTrack settings. Here {\large$\bullet$} indicates closed source models, and {\large$\blacktriangle$} indicates locally deployed models. Dashed arrows connect each model's BASE and LogicTrack results.}
\label{fig:overall_scatter}
\end{figure}

Figure~\ref{fig:overall_scatter} compares the average OA and VU of the base models and LogicTrack across all seven models. Overall, LogicTrack consistently improves both metrics on almost all settings. 
The VU gains are especially pronounced: GPT-4o-mini rises from 21.51\% to 46.61\%, Llama3.1-8B from 9.24\% to 32.75\%, and Qwen2.5-7B from 16.33\% to 35.54\%. Among the closed-source models, GPT-4o-mini also achieves the strongest joint gain in OA, increasing from 70.68\% to 78.49\%. The simultaneous improvement of OA and VWA suggests that verification does more than trigger backtracking; it can steer regeneration toward reasoning paths that are both more valid and more answer-preserving.

Table~\ref{tab:main_results} further confirms that this pattern holds at the benchmark level. 
LogicTrack improves over the baseline in 133 out of 168 benchmark metric comparisons.
Specifically, LogicTrack preserves or improves OA on most model and dataset pairs, with especially large gains on tasks that require multi-step deduction.
Qwen2.5-7B shows 0\% VWA on QASC as it directly outputs answers with no reasoning steps, causing step-level evaluation to fail. This issue can be mitigated by our SFT model (Table~\ref{tab:sft-all-models}).
GPT-4o-mini on ProofWriter rises from 54.27\% to 79.9\%, Qwen2.5-7B on ProntoQA from 92\% to 97\%, and Mistral-7B on ProofWriter from 27.14\% to 39.2\%. Even when the base model is already strong, LogicTrack maintains OA while improving verifiability. GPT-5-nano on ProntoQA, for example, stays at high OA while UR drops from 11.17\% to 10.24\%.
LogicTrack also reduces UR for most model and dataset pairs, confirming that backtracking replaces unverifiable steps with solver-validated alternatives. The largest reductions include Gemini-2.5 on FOLIO from 75.63\% to 27.79\%, Llama-3.1-8B on SARA from 39.1\% to 9.89\%, and Llama-3.1-8B on QASC from 44.65\% to 16.3\%. These reductions are accompanied by large VWA gains. GPT-4o-mini on ESNLI rises from 46.38\% to 87.21\%, and Gemini-2.5 on FOLIO from 21.17\% to 56.37\%. Overall, these results show that LogicTrack improves both final answers and the verifiability of the reasoning process.

\subsection{Ablation Study of LogicTrack}
\label{sec:ablation}

We perform an ablation study on SARA with both open-source and proprietary LLMs to examine the effects of different reward designs and model backtracking strategies on the performance of LogicTrack.

\textbf{Effectiveness of SBR reward.}
Table~\ref{tab:ablation-side-by-side}a ablates the contribution of each SBR component: fidelity reward only $R^{\mathrm{fidelity}}$, solver reward only $R^{\mathrm{verify}}$, and the full SBR reward.
The fidelity reward $R^{\mathrm{fidelity}}$ yields stronger outcome-based metric gains, with a $\Delta$OA of 10.7\% on GPT-4o-mini and 11\% on Llama-8B. Yet its effect on metrics that require step-level verification is far more modest. For instance, on the VWA metric, fidelity reward improves over the baseline by 5.4\% on GPT-4o-mini and a mere 0.4\% on Llama-8B. In contrast, the solver reward alone achieves VWA improvements exceeding 20\% on both models.
This suggests that grounding each step in the given premises helps the model avoid certain flawed steps and therefore contributes positively to final answer correctness. However, improving step-level verifiability still requires other reward signals. This gap supports the design choice of using solver feedback as a separate reward term.

This is validated by the results showing that reward $R^{\mathrm{verify}}$ greatly improved the VWA performance. On GPT-4o-mini, the solver reward alone achieves 23.4\% VWA and 12.9\% OA gains over the baseline, while reducing the unverified ratio from 32.55\% to 15.78\%. On Llama-8B, it delivers an identical $\Delta$VWA of $25.4\%$ and brings the unverified step ratio down from 39.10\% to 7.89\%. These results align with our expectation that the solver reward directly incentivizes the model to produce structurally valid, executable reasoning steps, which are precisely what VWA and UR measure.

Combining both rewards (SBR) yields the best VWA for both models with $\Delta$VWA of 25.4\% and 25.9\% over the baseline, as well as the lowest UR on GPT-4o-mini at 10.81\%. On Llama-8B, adding both rewards further lifts OA from 62.50\% to 63.60\%, confirming that these two signals are complementary. $R^{\mathrm{verify}}$ incentivizes structural correctness while $R^{\mathrm{fidelity}}$ enforces factual grounding, and together they cover both dimensions of high-quality structural logical reasoning.

\begin{table*}[t]
\centering
\begin{minipage}[t]{0.49\textwidth}
\vspace{0pt}
\centering
{\small (a) SBR Reward Component Ablation}\\[0.0em]
\resizebox{0.99\linewidth}{!}{%
\begin{tabular}{ccccccc}
\toprule
\textbf{Model} & \textbf{Method} & \textbf{VWA} $\Uparrow$ & $\Delta$\textbf{VWA} $\Uparrow$ & \textbf{OA} $\Uparrow$ & $\Delta$\textbf{OA} $\Uparrow$ & \textbf{UR} $\Downarrow$ \\
\midrule
\multirow{3}{*}{\shortstack[c]{gpt-\\4o\\mini}}
& Fidelity Only & 45.80 & 5.3  & 70.96 & 10.7 & 37.19 \\
& Solver Only   & 63.96 & 23.4 & 73.16 & 12.9 & 15.78 \\
& \ablationhlcell{SBR (Both)} & \ablationhlcell{\textbf{65.90}} & \ablationhlcell{\textbf{25.4}} & \ablationhlcell{73.16} & \ablationhlcell{12.9} & \ablationhlcell{\textbf{10.81}} \\
\midrule
\multirow{3}{*}{\shortstack[c]{Llama-\\3.1\\8B}}
& Fidelity Only & 32.49 & 0.4  & 63.97 & 11.0 & 48.83 \\
& Solver Only   & 57.55 & 25.4 & 62.50 & 9.6  & 7.89  \\
& \ablationhlcell{SBR (Both)} & \ablationhlcell{\textbf{57.98}} & \ablationhlcell{\textbf{25.9}} & \ablationhlcell{63.60} & \ablationhlcell{10.7} & \ablationhlcell{9.89} \\
\bottomrule
\end{tabular}
}
\end{minipage}
\hfill
\begin{minipage}[t]{0.49\textwidth}
\vspace{0pt}
\centering
{\small (b) Backtracking Search Strategy Variants}\\[0.0em]
\resizebox{0.94\linewidth}{!}{%
\begin{tabular}{ccccccc}
\toprule
\textbf{Model} & \textbf{Method} & \textbf{VWA} $\Uparrow$ & $\Delta$\textbf{VWA} $\Uparrow$ & \textbf{OA} $\Uparrow$ & $\Delta$\textbf{OA} $\Uparrow$ & \textbf{UR} $\Downarrow$ \\
\midrule
\multirow{3}{*}{\shortstack[c]{gpt-\\4o-\\mini}}
& Beam  & 54.87 & 14.4 & 69.12 & 8.8  & 23.79 \\
& MCTS  & 53.06 & 12.6 & 72.06 & 11.8 & 26.84 \\
& \ablationhlcell{Main} & \ablationhlcell{\textbf{65.90}} & \ablationhlcell{\textbf{25.4}} & \ablationhlcell{73.16} & \ablationhlcell{12.9} & \ablationhlcell{\textbf{10.81}} \\
\midrule
\multirow{3}{*}{\shortstack[c]{Llama-\\3.1-\\8B}}
& Beam & 56.71 & 24.6 & 65.44 & 12.5 & 12.88 \\
& MCTS & 56.19 & 24.1 & 64.34 & 11.4 & 13.54 \\
& \ablationhlcell{Main} & \ablationhlcell{\textbf{57.98}} & \ablationhlcell{\textbf{25.9}} & \ablationhlcell{63.60} & \ablationhlcell{10.7} & \ablationhlcell{9.89} \\
\bottomrule
\end{tabular}
}
\end{minipage}
\caption{Ablation Study of LogicTrack. (a) SBR reward component ablation. (b) Backtracking search strategy variants. The $\Delta$ values report the absolute improvement (\%) over the BASE model.}
\label{tab:ablation-side-by-side}
\end{table*}

\begin{figure*}[t]
  \centering
  \includegraphics[width=\linewidth, trim=0 10 0 10, clip]{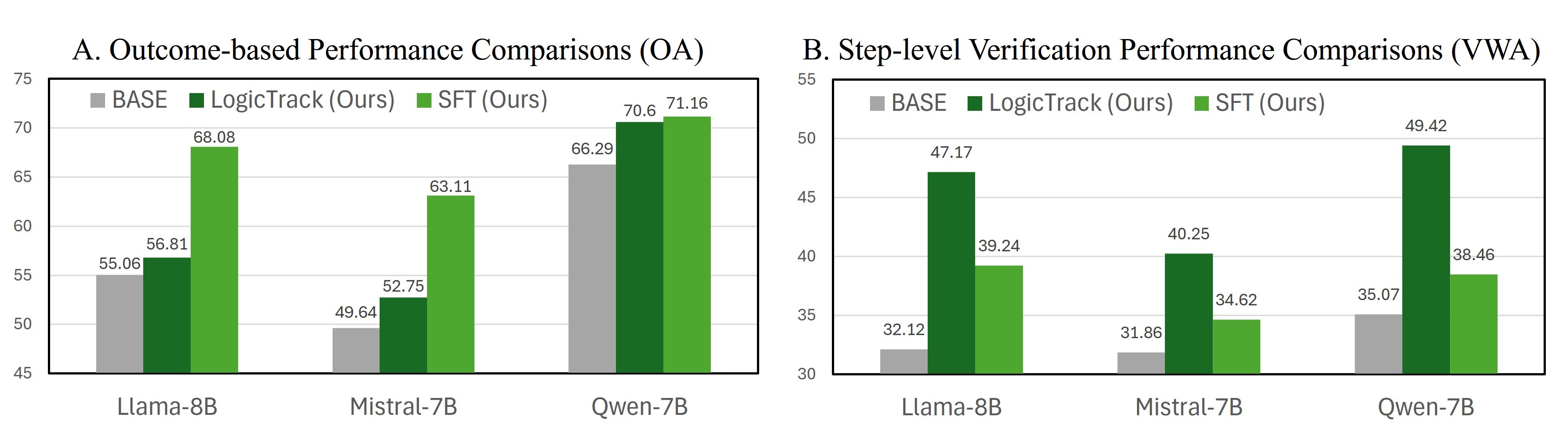}
  \caption{OA $\Uparrow$ and VWA $\Uparrow$ under BASE, LogicTrack, and LogicTrack-SFT averaged over all datasets. LogicTrack-SFT refers to the model fine-tuned on LogicTrack-generated train data.}
  \label{fig:sft_results}
\end{figure*}

\textbf{Backtracking Search Strategy Variants.}
Table~\ref{tab:ablation-side-by-side}b compares three search strategies, including default greedy backtracking, beam search, and Monte Carlo Tree Search.
The default strategy achieves the highest VWA and lowest UR across both models.
On GPT-4o-mini, it yields a $\Delta$VWA of 25.4\% and $\Delta$OA of 12.9\% over the baseline, with UR reduced to 10.81\%. Beam search and MCTS improve VWA by only 14.4\% and 12.6\% respectively and retain higher unverified ratios of 23.79\% and 26.84\%.
On Llama-8B, all three strategies converge in VWA improvement at 25.9\%, 24.6\%, and 24.1\%, yet the greedy strategy still achieves the lowest UR at 9.89\%, compared to 12.88\% and 13.54\%.

This advantage can be attributed to the greedy strategy's mechanism of immediate verification feedback: a failed step triggers the instant delivery of the solver's audit log and regeneration from the exact failure position, whereas beam search and MCTS distribute their search budget across parallel candidates. Notably, MCTS achieves a higher $\Delta$OA than beam search on GPT-4o-mini, at 11.8\% compared with 8.8\%. This suggests that broader exploration improves final-answer correctness, albeit at the expense of step-level verifiability.

\subsection{Extended Application of LogicTrack on SFT}
\label{sec:case_study}

LogicTrack operates at inference time to improve step-level 
verification of model reasoning. We further extend it to the construction of fine-tuning datasets, exploring its potential for producing self-backtracking SFT data.  
We first report overall performance after SFT, then present case studies that compare reasoning steps before and after backtracking to show the effectiveness of the fine-tuning.

\textbf{LogicTrack-SFT surpasses inference-time LogicTrack on answer correctness.}
As shown in Figure~\ref{fig:sft_results}a, models fine-tuned on LogicTrack backtracking trajectories achieve the highest OA across all models.
Compared with inference-time LogicTrack, SFT improves OA from 56.81\% to 68.08\% on Llama-8B, from 52.75\% to 63.11\% on Mistral-7B, and from 70.60\% to 71.16\% on Qwen-7B.
This suggests that training on solver-verified trajectories with explicit backtrack tokens enables the model to internalize reasoning patterns that generalize beyond what inference-time search alone achieves.

\textbf{SFT partially transfers step-level verification capability, though inference-time LogicTrack retains the advantage.}
Figure~\ref{fig:sft_results}b shows that SFT also improves step-level quality: VWA increases over the base model from 32.12\% to 39.24\% on Llama-8B, from 31.86\% to 34.62\% on Mistral-7B, and from 35.07\% to 38.46\% on Qwen-7B, showing that LogicTrack trajectories can transfer partial verification capability into the model's intrinsic reasoning.
However, inference-time LogicTrack still achieves the highest VWA, reaching 47.17\%, 40.25\%, and 49.42\% on the three models respectively.
This gap shows that keeping the solver in the loop at inference time provides stronger step-level guarantees. 

\textbf{Better performance than other state-of-the-art tuning methods.}
Monitoring reasoning trajectories remains an under-explored area, and the work most closely related to ours is LogicReward~\citep{xu2025logicreward}.
LogicReward designs a solver-based reward to score multiple responses for each problem and then uses the top responses to construct training data for SFT and DPO.
Table~\ref{tab:logicreward_logictrack_sft_overall_appendix} compares LogicReward with two variants of LogicTrack in terms of average performance across datasets.
The results show that our methods achieve stronger overall performance: inference-time LogicTrack improves OA from 0.4338 to 0.5765 and VWA from 0.2656 to 0.4782, while LogicTrack-SFT further achieves the best OA of 0.6808.
The full dataset-level results in Table~\ref{tab:logicreward_logictrack_sft_overall_appendix} also show that LogicTrack reduces UR from 0.4089 to 0.1863.
Since both LogicReward and LogicTrack-SFT rely on SFT, this improvement suggests that LogicTrack constructs higher-quality SFT training data.

\begin{figure}[t]
    \centering
    \includegraphics[width=\linewidth]{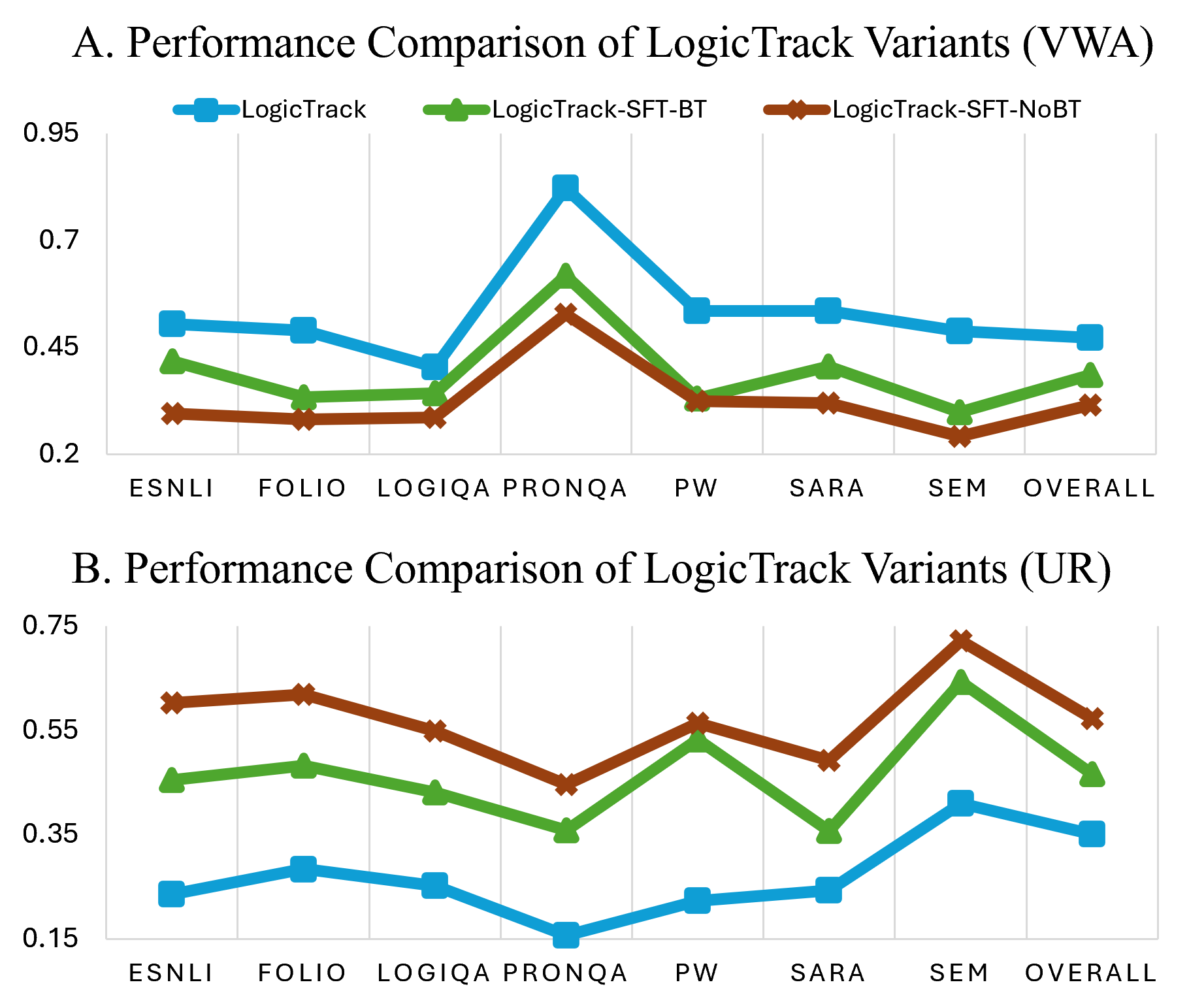}
    \caption{Effectiveness of backtracking supervision in LogicTrack-SFT on Qwen2.5-7B. The VWA $\uparrow$ and UR $\downarrow$ comparison across inference-time LogicTrack, LogicTrack-SFT-BT, and LogicTrack-SFT-NoBT.}
    \label{fig:cmp_nobt}
\end{figure}

To isolate the contribution of adding explicit backtracking supervision to training samples, we trained LogicTrack-SFT-NoBT, a control model trained on data without backtracking traces.
As shown in Figure~\ref{fig:cmp_nobt}, models fine-tuned with LogicTrack backtracking trajectory data consistently achieve higher VWA and lower UR than LogicTrack-SFT-NoBT across all benchmarks.
These results confirm that the observed gains are not only a generic effect of SFT on correct trajectories. The better performance of LogicTrack-SFT-BT over LogicTrack-SFT-NoBT demonstrates that incorporating backtracking trajectories provides additional supervision and can effectively enhance fine-tuned model performance.

\section{Related Work}
\label{sec:related}

\begin{table}[t]
\centering
\scriptsize
\setlength{\tabcolsep}{5pt}
\begin{tabular*}{\columnwidth}{@{\extracolsep{\fill}}lccc@{}}
\hline
\textbf{Metric} & \textbf{LogicReward}~\cite{xu2025logicreward} & \textbf{LogicTrack} & \textbf{LogicTrack-SFT} \\
\hline
\textbf{OA} $\uparrow$ & 0.4338 & 0.5765 & \textbf{0.6808} \\
\textbf{VWA} $\uparrow$ & 0.2656 & \textbf{0.4782} & 0.3924\\
\hline
\end{tabular*}
\caption{Performance comparison among LogicReward and LogicTrack. Full comparisons are in Table 6 of Appendix C.}
\label{tab:logicreward_logictrack_sft_overall_appendix}
\end{table}

\subsection{LLM Reasoning Capability Enhancement}
Many methods have been proposed to improve LLM reasoning patterns,
and they can be broadly categorized into training-based and training-free
approaches. 
Training-free approaches improve reasoning quality during inference without updating parameters, including prompting strategies~\citep{wang2022self, wei2022chain}, search-based methods~\citep{yang2025selfbacktracking, yao2023tree-of-thought}, and external knowledge augmentation~\citep{peng2023check}.
Training-based methods design rewards to identify high-quality reasoning data for fine-tuning~\citep{gulcehre2023reinforced,zelikman2022star}, or to incorporate these rewards into the training objective~\cite{luong2024reft}. Reward designs include outcome-based and process-level critics that attend to intermediate steps~\citep{yue2026promoting, calanzone2024logically, lightman2023lets, wang2024math}.
However, the criteria used to assess reasoning quality generally lack formal grounding. Outcome-based critics evaluate only final-answer correctness, while process-level critics rely on human annotations or LLM-generated judgments, limiting verifiability and allowing subtle logical errors to go undetected.
To address this, we define intermediate-step reasoning quality based on whether each step is logically supported by deterministic step-level feedback from solver verification.

\subsection{LLM Reasoning Chain Verification}
Recent work increasingly combines LLMs with verification mechanisms to improve soundness of model-generated reasoning~\citep{logic-survey1, logic-survey2}.
Existing approaches generally fall into two broad paradigms: formalizing LLM-generated responses and verifying them with external solvers~\citep{linc111, logic-lm}, or treating LLMs themselves as symbolic provers~\citep{xu2025aristotle, symbcot}.
This work focuses on external solver-based approaches, as they provide a more explicit and controllable verification signal than self-verification. Within this paradigm, \cite{zhang2025masa,ranaldi2025improving} discuss efficient auto-formalization; \cite{xu2026adaptive} study adaptive solver selection; \cite{quan-emnlp24, quan-acl25} use solver feedback for iterative refinement. These works mainly target reasoning tasks close to formal representations, or highly structured domains such as mathematics and programming~\citep{ren2025deepseek}, where complete responses can be directly passed to a solver for verification.
Step-level verification for general natural language reasoning remains under-explored, as it requires independently formalizing and checking each intermediate chain rather than one-off final answer verification.
In contrast, LogicTrack verifies the symbolic formalization induced from each NL reasoning step and backtracks at the first failed step, preventing local errors from propagating through the full reasoning chain.

\section{Conclusion}
\label{sec:conclusion}
We presented LogicTrack, a neuro-symbolic framework for generation-time auditing of LLM reasoning trajectories. LogicTrack decomposes each step into explicit context, explanations, and conclusions.
Following standard practice in solver-based verification, LogicTrack uses autoformalization as a modular interface that maps NL reasoning steps to solver-executable logical specifications.
Finally, LogicTrack uses the Solver-based Backtracking Reward (SBR) to combine premise fidelity with solver feedback for online backtracking and regeneration. Across eight reasoning benchmarks and seven LLMs, LogicTrack substantially improves step-level verifiability and generally preserves or improves final-answer accuracy, with ablations showing that solver verification drives most of the gains. LogicTrack backtracking traces also serve as useful supervision for fine-tuning open-source models, improving both final answer accuracy and verification performance over the base models. Future work on autoformalization optimazation, verification reward variants, backtracking strategies, and fine-tuning approaches that enable LLMs to internalize verification capabilities will be valuable.

\bibliography{aaai2027}

\clearpage
\appendix

\section{Appendix A: Methods Implementation}

\subsection{Backtracking Search Strategies}
\label{apx:backtrack-variant}
This section provides additional implementation details for the search-strategy ablation in Table~\ref{tab:ablation-side-by-side}(b).
We compare three variants: the default greedy backtracking search used in the main experiments, beam search, and Monte Carlo Tree Search (MCTS).

\paragraph{Default Search}

The default search is the greedy backtracking strategy used throughout the main experiments and the implementation is described in Section \ref{sec:tree_search}.
Its step-level acceptance threshold is set to $\theta_{\mathrm{step}} = 0.8$, the running-average threshold is set to $\theta_{\mathrm{avg}} = 0.5$. The maximum number of regeneration attempts per step position is $k = 2$. If all $k$ attempts fail, the candidate with the highest $\mathrm{SBR}$ is force-forwarded to guarantee forward progress.

\paragraph{Beam Search}

Beam search maintains a frontier of $B$ partial reasoning paths and expands all of them in parallel at each layer.
At each layer, every frontier path contributes up to $k=2$ continuation branches: if the path carries a pending tail from a prior generation, its leading provisional step is consumed as one free branch; the remaining slots are filled by fresh LLM completions conditioned on the current verified prefix.
All verification tasks are dispatched concurrently.
After verification, nodes with $\mathrm{SBR}(s_t) < 0.75$ are pruned; the top-$B$ survivors ranked by path-average $\mathrm{SBR}$ advance to the next layer.
If all candidates are pruned, the best low-reward node is reactivated to guarantee forward progress.
The search terminates as soon as the top frontier node yields a complete reasoning chain; otherwise the best node found is returned.

\paragraph{Monte Carlo Tree Search}

We follow the standard MCTS framework, repeating four stages each iteration. During selection, the algorithm traverses from the root by choosing at each node the child with the highest UCB score, which balances the mean backpropagated $\mathrm{SBR}$ against an exploration bonus, until it reaches a node with fewer than $k$ children. During expansion, up to $k$ fresh LLM completions are generated from the selected node. The first step $s_t$ of each completion is verified and attached as a new child. A child is marked terminal if $\mathrm{SBR}(s_t)=0$, if the reasoning chain is complete, or if a depth limit is reached. During rollout, each non-terminal child whose $\mathrm{SBR}(s_t)\geq\theta$ is extended by consuming the remaining provisional steps from the same generation, verifying each in turn without additional LLM calls, until the chain terminates or a step falls below $\theta$. During backpropagation, the path-average $\mathrm{SBR}$ of the reached leaf is propagated back to the root, updating each ancestor's statistics. After a fixed number of iterations the best complete leaf is returned, or, if none exists, the best leaf overall.

\subsection{Backtracking SFT}
\label{apx:backtrack-sft}

We build the SFT data from backtracking traces produced by LogicTrack.
For each training example, we keep only cases whose final answer is correct and whose reasoning steps pass both syntactic and semantic verification.
From each trace, we extract the verified path $\mathbf{s}^{+} = (s_1^{+}, \ldots, s_n^{+})$ from \texttt{final\_path} and the rejected steps $\{s_k^{-}\}$ collected at backtracking points.
A rejected step at position $k$ is kept only when its prefix up to step $k-1$ matches the verified path.
This keeps the error local and avoids trajectories that have already diverged earlier.

We then construct three kinds of training trajectories:
\begin{itemize}
\item \textbf{Single backtracking correction.}
One wrong step is followed by a dedicated backtracking token and then the verified continuation:
\[
  s_1^{+} \cdots s_{k-1}^{+} \;\; s_k^{-} \;\; \texttt{<backtrack>} \;\; s_k^{+} \cdots s_n^{+} \;\; \boxed{\hat{y}}.
\]
\item \textbf{Multiple backtracking corrections.}
These trajectories contain two or more corrections at different step positions. We prefer cases whose wrong continuation ends with an incorrect final answer.

\item \textbf{Clean correct trajectories.}
These trajectories contain all verified steps and provide positive examples.

\end{itemize}

During fine-tuning, we add \texttt{<backtrack>} as the new vocabulary item.
The model learns to emit this token when a step is wrong and then continue with a corrected path, without relying on an external verifier at inference time.
The final training set contains $~\sim$6{,}000 samples across the three trajectory types.
We fine-tune three large language models on the constructed SFT dataset: Qwen2.5-7B-Instruct, Mistral-7B-Instruct-v0.3, and Llama-3.1-8B-Instruct.
We use LoRA supervised fine-tuning for 2 epochs with rank 32 and scaling factor 64.
Training uses bfloat16 and AdamW (\texttt{adamw\_torch}) with a learning rate of $3 \times 10^{-5}$, a linear scheduler, warmup ratio 0.05, and gradient clipping with maximum norm 0.3.
The per-device batch size is 2 and the effective batch size is 16.
We also enable gradient checkpointing and save checkpoints every 50 steps.

\section{Appendix B: Experimental Settings}

\subsection{Datasets}
\label{app:datasets}

We evaluate on eight reasoning benchmarks.
For e-SNLI, FOLIO, LogiQA, ProntoQA, ProofWriter, and QASC, we adopt the processed releases used in LogicReward~\citep{xu2025logicreward}.
We then resplit their released `data.jsonl` into training and test subsets with the ratio of 4 : 1. All fine-tuning experiments are based on the training dataset.
For SemEval~\citep{semeval2026task11}, Task 2 is used for testing and the English portion of Task 4 is used for training.
For SARA~\citep{holzenberger2020dataset}, we follow the instructions from \cite{feng2025vericot} and use the instances from SARA Entailment in the testing stage only.
We briefly summarize the benchmarks below.

\begin{itemize}
    \item \textbf{e-SNLI}~\citep{camburu2018esnli} extends SNLI with human-annotated natural language explanations for premise--hypothesis pairs.

    \item \textbf{FOLIO}~\citep{han2024folio} is a human-annotated benchmark for natural language reasoning with accompanying first-order logic annotations that are automatically verified by an inference engine.

    \item \textbf{LogiQA}~\citep{liu2020logiqa} is originally a multiple-choice reading comprehension benchmark for logical reasoning over short passages. We use the preprocessed version from \cite{xu2025logicreward}.

    \item \textbf{ProntoQA}~\citep{prontoqaRef} is a synthetic question answering benchmark in which each example is generated from a synthetic world model represented in first-order logic, providing a controlled setting for analyzing multi-step deduction.

    \item \textbf{ProofWriter}~\citep{tafjord2021proofwriter} is a synthetic benchmark over natural-language theories of facts and rules, with associated questions and proofs at different reasoning depths.

    \item \textbf{QASC}~\citep{khot2020qasc} is an 8-way multiple-choice science question answering benchmark for question answering via sentence composition.

    \item \textbf{SARA}~\citep{holzenberger2020dataset} is a benchmark for statutory reasoning in tax law, with entailment and question answering tasks. 

    \item \textbf{SemEval}~\citep{semeval2026task11} refers to SemEval-2026 Task 11 on disentangling content from formal reasoning in multilingual syllogistic arguments. We use Task 2 for testing and the English portion of Task 4 for training, and in both cases use the binary validity labels.

\end{itemize}

\subsection{Models}
\label{app:models}

Our experiments cover seven models with diverse architectures and scales. 

\paragraph{Proprietary models.}
\begin{itemize}
    \item \textbf{GPT-4o-mini}~\citep{hurst2024gpt4o} is a member of the GPT-4o family that offers strong reasoning performance at low inference cost.
    \item \textbf{GPT-5-nano}~\citep{openai2025gpt5} is the smallest model in the GPT-5 family and is designed for fast and lightweight inference.
    \item \textbf{Gemini-2.5-Flash-Lite}~\citep{comanici2025gemini} is a lightweight model from the Gemini 2.5 family that balances efficiency with strong reasoning ability.
\end{itemize}

\paragraph{Open weight models.}
\begin{itemize}
    \item \textbf{Llama-3.1-8B-Instruct}~\citep{llama3} is an instruction-tuned model with 8B parameters from Meta's Llama 3.1 family.
    \item \textbf{Mistral-7B-Instruct-v0.3}~\citep{jiang2023mistral7b} is an instruction-tuned model with 7B parameters from Mistral AI with strong performance relative to its size.
    \item \textbf{Qwen2.5-7B-Instruct}~\citep{yang2024qwen2} is an instruction-tuned model with 7B parameters from the Qwen2.5 family.
    \item \textbf{Qwen2.5-14B-Instruct}~\citep{yang2024qwen2} is an instruction-tuned model with 14B parameters from the same Qwen2.5 family with stronger reasoning capacity at higher computing cost.
\end{itemize}

\paragraph{Implementations.} 
For all models, the temperature is set to 0.7 and the maximum generation length is set to 10240 tokens when answering questions.
For proprietary models, experiments are performed via corresponding OpenAI APIs\footnote{\url{https://developers.openai.com/api/docs}} and Gemini APIs\footnote{\url{https://ai.google.dev/gemini-api/docs}}. For open-weight models, all testing and fine-tuning experiments are conducted on NVIDIA A100 GPUs (40 GB or 80 GB) or equivalent GPUs, using PyTorch 2.8 and Python 3.12.
The BASE method in this paper is the zero-shot strategy that uses the Reasoning Prompt described in Appendix D. 
Our LogicTrack method uses the same reasoning prompt as the baseline and applies our monitoring mechanism during answer generation. For formal verification, we use the `gpt-4o-min' as the autoformalizer backend and Z3 SMT solver as the solver backend. The automatic formalizer and fidelity judge in LogicTrack are both implemented with gpt-4o-mini.
Evaluating the intermediate reasoning processes of LLMs remains an underexplored area. To the best of our knowledge, LogicReward \cite{xu2025logicreward} is the only relevant prior work with open-source code. We select it as baseline and the performance comparison between LogicTrack and LogicReward is presented in Appendix C.

\subsection{Metric Definitions}
\label{app:metrics}

For each problem $x$, let $T_x$ be the number of reasoning steps and $K_x$ the number of verified steps, giving the per-problem verification ratio $\rho_x = K_x / T_x$.

Outcome Accuracy (OA) is the fraction of examples with correct final answers.
Verified Ratio (VR) is the dataset-average verification ratio $\tfrac{1}{N}\sum_x \rho_x$, Unverified Ratio (UR) is the mean fraction of unverified steps $\tfrac{1}{N}\sum_x (1-\rho_x)$.

Verified Utility (VU) is the fraction of all examples for which the final answer is correct and the entire reasoning trajectory is verified: $\mathrm{VU} = \frac{1}{N} \sum_{x=1}^N \mathbf{1}[\hat{y}_x = y_x] \cdot \mathbf{1}[\rho_x = 1] $.

Verification-Weighted Accuracy (VWA) is defined as $\tfrac{1}{N}\sum_{x=1}^N \rho_x \cdot \mathbf{1}[\hat{y}_x = y_x]$, it weights each correct prediction by its per-problem verification ratio.

\section{Appendix C: Results}

\subsection{More Results of LogicTrack Performance}
Table~\ref{tab:overall_performance_appendix} reports the average OA and VU of the base models and LogicTrack across all seven models.
Table~\ref{tab:sft-all-models} reports the performance of the three LogicTrack-SFT models on each dataset.
We further compare LogicTrack with LogicReward~\cite{xu2025logicreward} on Llama-8B.
LogicReward is based on their released model after SFT and DPO fine-tuning\footnote{\url{https://huggingface.co/Aiden0526/LogicReward-Llama3.1-8B}}.
Table~\ref{tab:logicreward_logictrack_sft_dataset_appendix} compares LogicReward, LogicTrack, and LogicTrack-SFT at the dataset level.

\begin{table*}[http]
\centering
\scriptsize
\setlength{\tabcolsep}{4pt}
\resizebox{\textwidth}{!}{
\begin{tabular}{@{}l*{14}{c}@{}}
\toprule
\multirow{2}{*}{\textbf{Metric}} & \multicolumn{2}{c}{\textbf{Gemini-2.5}} & \multicolumn{2}{c}{\textbf{GPT-4o-mini}} & \multicolumn{2}{c}{\textbf{GPT-5-nano}} & \multicolumn{2}{c}{\textbf{Llama-3.1-8B}} & \multicolumn{2}{c}{\textbf{Mistral-7B}} & \multicolumn{2}{c}{\textbf{Qwen2.5-7B}} & \multicolumn{2}{c}{\textbf{Qwen2.5-14B}} \\
\cmidrule(lr){2-3}\cmidrule(lr){4-5}\cmidrule(lr){6-7}\cmidrule(lr){8-9}\cmidrule(lr){10-11}\cmidrule(lr){12-13}\cmidrule(lr){14-15}
 & \textbf{Base} & \textbf{LogicTrack} & \textbf{Base} & \textbf{LogicTrack} & \textbf{Base} & \textbf{LogicTrack} & \textbf{Base} & \textbf{LogicTrack} & \textbf{Base} & \textbf{LogicTrack} & \textbf{Base} & \textbf{LogicTrack} & \textbf{Base} & \textbf{LogicTrack} \\
\midrule
\textbf{OA} $\Uparrow$ & 0.8096 & 0.8112 & 0.7068 & 0.7849 & 0.6850 & 0.7044 & 0.5506 & 0.5681 & 0.4964 & 0.5275 & 0.6629 & 0.7060 & 0.7402 & 0.7506 \\
\textbf{VU} $\Uparrow$ & 0.1936 & 0.2725 & 0.2151 & 0.4661 & 0.2607 & 0.3968 & 0.0924 & 0.3275 & 0.1841 & 0.2980 & 0.1633 & 0.3554 & 0.2382 & 0.4143 \\
\textbf{VWA} $\Uparrow$ & 0.3071 & 0.3693 & 0.4578 & 0.6436 & 0.4527 & 0.5394 & 0.3212 & 0.4717 & 0.3186 & 0.4025 & 0.3507 & 0.4942 & 0.4484 & 0.5573 \\
\textbf{UR} $\Downarrow$ & 0.6273 & 0.5477 & 0.3727 & 0.1898 & 0.3832 & 0.2570 & 0.4171 & 0.1871 & 0.3793 & 0.2476 & 0.4737 & 0.3068 & 0.4213 & 0.2687 \\
\bottomrule
\end{tabular}
}
\caption{Overall performance for all models averaged across eight benchmarks under BASE and LogicTrack. Here OA and VU correspond to the coordinates used in Figure~\ref{fig:overall_scatter}}
\label{tab:overall_performance_appendix}
\end{table*}

\begin{table*}[htbp]
\centering
\scriptsize
\setlength{\tabcolsep}{4pt}
\resizebox{\textwidth}{!}{
\begin{tabular}{@{}cc*{8}{c}@{}}
\toprule
\textbf{Model}
& \textbf{Metric}
& \textbf{ESNLI}
& \textbf{FOLIO}
& \textbf{LogiQA}
& \textbf{ProntoQA}
& \textbf{ProofWriter}
& \textbf{QASC}
& \textbf{SARA}
& \textbf{SemEval} \\
\midrule

\multirow{3}{*}{\shortstack{SFT \\ \\ Llama-3.1-8B}}
& \textbf{OA}
& 0.6047 & 0.6121 & 0.5170 & 0.9900
& 0.6548 & 0.8767 & 0.6324 & 0.7842 \\
& \textbf{VWA}
& 0.3366 & 0.3412 & 0.2922 & 0.8069
& 0.3654 & 0.5194 & 0.4029 & 0.3004 \\
& \textbf{UR}
& 0.4012 & 0.4735 & 0.4413 & 0.1856
& 0.5069 & 0.4463 & 0.3836 & 0.6207 \\
\midrule

\multirow{3}{*}{\shortstack{SFT \\ \\Qwen2.5-7B}}
& \textbf{OA}
& 0.7984 & 0.6552 & 0.6136 & 0.9700
& 0.6131 & 0.8767 & 0.6360 & 0.7895 \\
& \textbf{VWA}
& 0.4180 & 0.3329 & 0.3428 & 0.6183
& 0.3302 & 0.4840 & 0.4053 & 0.2982 \\
& \textbf{UR}
& 0.4554 & 0.4825 & 0.4313 & 0.3587
& 0.5335 & 0.4589 & 0.3580 & 0.6433 \\
\midrule
\multirow{3}{*}{\shortstack{SFT \\ \\ Mistral-7B}}
& \textbf{OA}
& 0.4651 & 0.5862 & 0.4943 & 0.9800
& 0.6432 & 0.8904 & 0.5184 & 0.7632 \\
& \textbf{VWA}
& 0.1970 & 0.3099 & 0.2846 & 0.7046
& 0.3566 & 0.5057 & 0.3232 & 0.2991 \\
& \textbf{UR}
& 0.4864 & 0.5037 & 0.4380 & 0.2934
& 0.5216 & 0.4372 & 0.3958 & 0.6246 \\

\bottomrule
\end{tabular}
}
\caption{Performance of LogicTrack-SFT across different models and datasets.
We report OA $\uparrow$, VWA $\uparrow$, and UR $\downarrow$.}
\label{tab:sft-all-models}
\end{table*}

\begin{table*}[htbp]
\centering
\scriptsize
\setlength{\tabcolsep}{3pt}
\resizebox{\textwidth}{!}{
\begin{tabular}{@{}lccccccccc@{}}
\toprule
\multirow{2}{*}{\textbf{Dataset}}
& \multicolumn{3}{c}{\textbf{OA} $\uparrow$}
& \multicolumn{3}{c}{\textbf{VWA} $\uparrow$}
& \multicolumn{3}{c}{\textbf{UR} $\downarrow$} \\
\cmidrule(lr){2-4}\cmidrule(lr){5-7}\cmidrule(lr){8-10}
& \textbf{LogicReward} & \textbf{LogicTrack} & \textbf{LogicTrack-SFT}
& \textbf{LogicReward} & \textbf{LogicTrack} & \textbf{LogicTrack-SFT}
& \textbf{LogicReward} & \textbf{LogicTrack} & \textbf{LogicTrack-SFT} \\
\midrule
FOLIO & 0.3103 & 0.4569 & 0.6121 & 0.1841 & 0.3940 & 0.3412 & 0.4490 & 0.1745 & 0.4735 \\
ESNLI & 0.0155 & 0.2171 & 0.6047 & 0.0052 & 0.1907 & 0.3366 & 0.2605 & 0.1118 & 0.4012 \\
LogiQA & 0.3920 & 0.5000 & 0.5170 & 0.2408 & 0.4472 & 0.2922 & 0.3655 & 0.1247 & 0.4413 \\
ProntoQA & 0.6100 & 0.8700 & 0.9900 & 0.4457 & 0.7152 & 0.8069 & 0.3085 & 0.2046 & 0.1856 \\
ProofWriter & 0.4121 & 0.5729 & 0.6548 & 0.2130 & 0.4291 & 0.3654 & 0.5150 & 0.2910 & 0.5069 \\
QASC & 0.6027 & 0.7534 & 0.8767 & 0.3144 & 0.6402 & 0.5194 & 0.4685 & 0.1630 & 0.4463 \\
SemEval & 0.5105 & 0.6053 & 0.7842 & 0.3222 & 0.4293 & 0.3004 & 0.5778 & 0.3215 & 0.6207 \\
SARA & 0.6176 & 0.6360 & 0.6324 & 0.3996 & 0.5798 & 0.4029 & 0.3263 & 0.0989 & 0.3836 \\
\bottomrule
\end{tabular}
}
\caption{Dataset-level comparison among LogicReward, LogicTrack, and LogicTrack-SFT on Llama-3.1-8B. We report OA $\uparrow$, VWA $\uparrow$, and UR $\downarrow$.}
\label{tab:logicreward_logictrack_sft_dataset_appendix}
\end{table*}

\subsection{Backtracking Case Study in LogicTrack-SFT}
\label{app:example107}

Figure~\ref{fig:case_study} presents two representative examples of self-backtracking from LogicTrack-SFT (Qwen2.5-7B). 
In Example~1, the model makes a non-sequitur by deriving an unrelated property instead of continuing the type chain. After backtracking, it completes the correct deduction. Example~2 shows an irrelevant derivation in which the model explores an off-topic property. The model triggers backtracking in these cases and, after backtracking, follows the targeted reasoning path.
These cases highlight LogicTrack-SFT's internalized backtracking ability and show how it recovers from non-sequitur inferences and unproductive reasoning paths in a way that is consistent with the solver-guided corrections in LogicTrack training trajectories. 

\par\medskip
\begingroup
\small
\noindent\textbf{Example 1: Non-sequitur inference.}
\textit{Premises:} Alex is a tumpus $\Rightarrow$ jompus $\Rightarrow$ dumpus $\Rightarrow$ zumpus $\Rightarrow$ yumpus; each yumpus is not cold. \textit{Query:} Alex is cold. Answer label: \textsc{False}\\[3pt]
\begin{tcolorbox}[colback=white, colframe=black!60, title={\scriptsize Before \texttt{<backtrack>}}, fonttitle=\bfseries, boxrule=0.4pt, left=3pt, right=3pt, top=2pt, bottom=2pt]
\footnotesize
\textbf{Step 4:} $\{C_4,E_4,q_4\}$, where $q_4:$ ``Alex is not a wumpus (since zumpuses are dull and cannot be cold).''\\[1pt]
--------------

\textit{$\times$ Non-sequitur: ``dull'' is unrelated to ``cold''; the chain must reach yumpus.}
\end{tcolorbox}

\begin{tcolorbox}[colback=white, colframe=black!60, title={\scriptsize After \texttt{<backtrack>}}, fonttitle=\bfseries, boxrule=0.4pt, left=3pt, right=3pt, top=2pt, bottom=2pt]
\footnotesize
\textbf{Step 4:} ``Alex is a yumpus (every zumpus is a yumpus).''\\
\textbf{Step 5:} ``Each yumpus is not cold $\Rightarrow$ Alex is not cold.''\\[1pt]
--------------

\textit{$\checkmark$ Answer: \textbf{False}.}
\end{tcolorbox}
\vspace{6pt}
\noindent\textbf{Example 2: Irrelevant derivation.}
\textit{Premises:} Sally is a yumpus $\Rightarrow$ zumpus $\Rightarrow$ numpus; each numpus is not bitter. \textit{Query:} Sally is not bitter. Answer label \textsc{True}\\[3pt]
\begin{tcolorbox}[colback=white, colframe=black!60, title={\scriptsize Before \texttt{<backtrack>}}, fonttitle=\bfseries, boxrule=0.4pt, left=3pt, right=3pt, top=2pt, bottom=2pt]
\footnotesize
\textbf{Step 1:} $\{C_1,E_1,q_1\}$, where $q_1:$``Sally is a yumpus, which means she is not opaque. But we need to determine if she is bitter or not.''\\[1pt]
--------------

\textit{$\times$ Derives irrelevant property (``not opaque'').}
\end{tcolorbox}

\begin{tcolorbox}[colback=white, colframe=black!60, title={\scriptsize After \texttt{<backtrack>}}, fonttitle=\bfseries, boxrule=0.4pt, left=3pt, right=3pt, top=2pt, bottom=2pt]
\footnotesize
\textbf{Step 1:} ``Sally is a yumpus $\Rightarrow$ zumpus $\Rightarrow$ numpus.''\\
\textbf{Step 2:} ``Each numpus is not bitter $\Rightarrow$ Sally is not bitter.''\\[1pt]
--------------

\textit{$\checkmark$ Answer: \textbf{True}.}
\end{tcolorbox}
\captionof{figure}{Self-backtracking Examples of LogicTrack-SFT. Each pair contrasts the flawed step with the corrected continuation. }
\label{fig:case_study}
\endgroup
\par\medskip


The examples above displayed the traces in the simplified way. Below displays full settings of Example~2 in Figure~\ref{fig:case_study}. It shows the original flawed step that triggers \texttt{<backtrack>} and the regenerated continuation.

\paragraph{Original Given premises.} P=' Every yumpus is not opaque; Yumpuses are zumpuses; Zumpuses are nervous; Every zumpus is a tumpus; Each tumpus is not large; Tumpuses are impuses; Impuses are temperate; Impuses are numpuses; Every wumpus is bitter; Each numpus is not bitter; Numpuses are vumpuses; Every vumpus is not kind; Vumpuses are rompuses; Rompuses are brown; Rompuses are jompuses; Every jompus is wooden; Every jompus is a dumpus; Sally is a yumpus.'
\textbf{Query:} Sally is not bitter. \hfill \textbf{Answer label:} \textsc{True}

Model generated answer with backtracking is as follows.
\newpage
\paragraph{Initial flawed step and backtracking trigger.}
\begin{lstlisting}
Step 1:
##Premises:## Sally is a yumpus; Every yumpus is not opaque; Yumpuses are zumpuses; Zumpuses are nervous; Every zumpus is a tumpus; Each tumpus is not large; Tumpuses are impuses; Impuses are temperate; Impuses are numpuses; Every wumpus is bitter; Each numpus is not bitter; Numpuses are vumpuses; Every vumpus is not kind; Vumpuses are rompuses; Rompuses are brown; Rompuses are jompuses; Every jompus is wooden; Every jompus is a dumpus.
##Derived Fact:## Sally is a yumpus, which means she is not opaque. However, we need to determine if she is bitter or not.
<backtrack>
Step 1:
##Premises:## Sally is a yumpus; Every yumpus is not opaque; Yumpuses are zumpuses; Zumpuses are nervous; Every zumpus is a tumpus; Each tumpus is not large; Tumpuses are impuses; Impuses are temperate; Impuses are numpuses; Every wumpus is bitter; Each numpus is not bitter; Numpuses are vumpuses; Every vumpus is not kind; Vumpuses are rompuses; Rompuses are brown; Rompuses are jompuses; Every jompus is wooden; Every jompus is a dumpus.
##Derived Fact:## Sally is a yumpus. Since yumpuses are zumpuses and zumpuses are numpuses, we can conclude that Sally is also a numpus.
Step 2:
##Premises:## Sally is a numpus; Each numpus is not bitter.
##Derived Fact:## Sally is not bitter.
Step 3:
##Premises:## The hypothesis states: Sally is not bitter; We have established that Sally is not bitter.
##Derived Fact:## The hypothesis matches the derived fact above. Therefore, the claim is true. The final answer is \boxed{True}.
\end{lstlisting}

\section{Appendix D: Prompt Templates}
\label{app:prompts}

This section includes the prompt templates used in our experiments, including the reasoning prompt, the consistency check prompt, and the answer extraction procedure.

\newpage
\paragraph{Reasoning Prompt}
\label{app:prompt_reasoning}

Given a sample $x$, the reasoning model receives a system message instructing it to reason step-by-step in the structured format described in \S~\ref{sec:formalization}, followed by a user message containing the premises $P$, query $Q$, and a dataset-specific answer instruction.
The system prompt includes an in-context demonstration and enforces the three-part step format (Premises, Explanation, Derived Fact).

\begin{tcolorbox}[title=System Prompt for Step-by-Step Reasoning]
\begin{lstlisting}
You are a meticulous logician. Read the user's given premise and hypothesis carefully, then reason step by step explicitly in a Natural Language Inference (NLI) style, and conclude with \boxed{LABEL}.
Do not stop reasoning just because the current facts do not prove the claim. Stop only when the claim is settled, or when no unresolved inference path can still settle it.

Each step in the response stage must follow the format below, which contains three sections: Premises, Assumptions, and Derived Fact:

**Response Format**
Step X:
##Premises:##
- [Facts from the input premises or prior derived facts.]
- [Don't include any user query or hypothesis here.]
##Explanation:##
- [Add linguistic bridges only (e.g., synonyms, lexical equivalences, definitional mappings)]
- [Never use Assumptions to fill logical gaps. Write None if no assumption is needed]
##Derived Fact:##
- [The derived fact must be fully supported by that step's premises and assumptions.]
- [Do not speculate about other facts.]

After your final step, output exactly your answer with \boxed{LABEL}.


\end{lstlisting}
\end{tcolorbox}


\paragraph{Consistency Check Prompt}
\label{app:prompt_consistency}

The consistency check described in \S~\ref{sec:verification} uses an LLM judge to determine whether the explanations $E_i$ and context $C_i$ conflict with the original given premises $P$. The judge is instructed to be lenient, only flag direct and unambiguous conflicts.

\begin{tcolorbox}[title=System Prompt for Consistency Check]
\begin{lstlisting}
You are a logical consistency checker. Determine whether the assumption in the reasoning step conflicts with the overall context.

*Examples*
- Consistent: given context contains 'the infant is crying, Tom is round', and the assumption says 'infants are babies; given context includes Tom is round'.
- Inconsistent: given context contains 'dog is round, cat is green', and the assumption says 'we don't know if cat is round'. Because it introduces information not grounded in the current context.

Respond with a JSON object in this exact format:
{"check_result": "consistent", "reason": "..."}
or
{"check_result": "inconsistent", "reason": "..."}
\end{lstlisting}
\end{tcolorbox}

\paragraph{Answer Extraction}
\label{app:answer_extraction}

The predicted answer $\hat{y}$ is extracted from the model's response via a rule-based procedure. We first look for the \verb|\boxed{...}| notation specified in the prompt, extracting the innermost content and stripping wrappers (e.g., \verb|\text{}|). If no boxed answer is found, we also consider alternative patterns such as \verb|<answer>...</answer>| tags and common answer-declaration phrases (e.g., ``the answer is \{LABEL\}'') in the final sentences of the response. If none of these patterns match, the prediction is marked as unanswered.

\end{document}